# Data Analysis with Bayesian Networks: A Bootstrap Approach

**Nir Friedman**
The Institute of Computer Science
The Hebrew University
Jerusalem 91904, ISRAEL
nir@cs.huji.ac.il

**Moises Goldszmidt**
SRI International
333 Ravenswood Ave.
Menlo Park, CA 94025
moises@erg.sri.com

**Abraham Wyner**
Department of Statistics, Wharton School
University of Pennsylvania
Philadelphia, PA
ajw@stat.wharton.upenn.edu

## Abstract

In recent years there has been significant progress in algorithms and methods for inducing Bayesian networks from data. However, in complex data analysis problems, we need to go beyond being satisfied with inducing networks with high scores. We need to provide confidence measures on features of these networks: Is the existence of an edge between two nodes warranted? Is the Markov blanket of a given node robust? Can we say something about the ordering of the variables? We should be able to address these questions, even when the amount of data is not enough to induce a high scoring network. In this paper we propose Efron's Bootstrap as a computationally efficient approach for answering these questions. In addition, we propose to use these confidence measures to induce better structures from the data, and to detect the presence of latent variables.

## 1 Introduction

In the last decade there has been a great deal of research focused on learning Bayesian networks from data [2, 12]. With few exceptions, these results have concentrated on computationally efficient induction methods and, more recently, on the issue of hidden variables and missing data. The main concern in this line of work is the induction of high *scoring* networks, where the score of the network reflects how well does the network fits the data. A Bayesian network, however, also contains structural and qualitative information about the domain. We should be able to exploit this information in complex data analysis problems, even in situations where the available data is sparse.

Part of our motivation comes from our ongoing work on an application of Bayesian networks to molecular biology [11]. One of the central goals of molecular biology is to understand the mechanisms that control and regulate gene expression. A gene is *expressed* via a process that *transcribes* it into an RNA sequence, and this RNA sequence is in turn *translated* into a protein molecule. Recent technical breakthroughs in molecular biology enable biologists to measure of the expression levels of thousands of genes in one experiment [6, 17, 21]. The data generated from these experiments consists of instances, each one of which has thousands of attributes. However, the largest datasets available today contain only few hundreds of instances. We cannot expect to learn a detailed model from such a sparse data set. However, these data sets clearly contain valuable information. For example, we would like to induce correlation and causation relations among genes (e.g., high expression levels of one gene "cause" the suppression of another) [16]. The challenge is then, to separate the measurable "signal" in this data from the "noise," that is, the genuine correlations and causations properties from spurious (random) correlations.

Analysis of such data poses many challenges. In this paper we examine how we can determine the level of confidence about various structural features of the Bayesian networks we induce from data sets. We consider an approach and methodology based on the Bootstrap method of Efron [7] for addressing this type of challenges. The Bootstrap is a computer-based method for assigning measures of accuracy to statistics estimates and performing statistical inference. We regard these measures of accuracy as establishing a level of *confidence* on the estimates, where confidence can be interpreted in two ways. The more important (and more elusive) notion assesses the likelihood that a given feature is actually true. This confidence will, ultimately, stand or fall by the method of estimation. The second notion is more akin to an assessment of the degree of support of a *particular technique* towards a given feature. This latter idea nicely separates the variation in the data from the shortcomings of the algorithm. It is this latter interpretation of confidence that was pursued in [10]. The methods introduced in this paper encompass both types of confidence, and focuses on the former (more below).

Although the Bootstrap is conceptually easy to implement and apply in our context, there are open question in the theoretical foundations. The main difficulty (as compared to classic statistical estimation methods) is the lack of closed form expressions for the events under study (e.g., that an edge appears in a network). Still, the widespread use of the bootstrap despite such difficulties reflects the general conditions under which bootstrap distributions are consistent, even when the statistics cannot be concisely defined in a simple expression (see [7]). An example is the application of the bootstrap in evolutionary biology to measure confidence in inferences from phylogenetic trees. Felsenstein [9], has applied re-sampling tools to estimate uncertainty in edges (clades) of evolutionary trees (which specify the phylogenetic evolution of a gene over time).

Similar to phylogenies, we test re-sampling strategies for Bayesian networks, experimentally, by beginning with an explicit probability distribution and a known network model (the "golden model"). In [10], we report preliminary results that indicate that, in practice, high confidence estimates on certain structural features are indicative of the existence of these features in the generating model. In these experiments, we used edges (clades) in *partially directed graphs* (PDAGs) as the feature of interest. These edges describe features of equivalence classes of networks (see below).

This paper extends the results in [10] in three fundamental ways: First it includes other important features of the induced models such as the Markov neighborhood of a node (i.e., with what confidence can we assert that $X$ is in $Y$'s Markov Blanket), and ordering relations between variables in the PDAGS (with what confidence can we assert that $X$ is an ancestor of $Y$). Second, we focus on examining to what extend the degree of confidence returned by the bootstrap can be interpreted as establishing the likelihood of a feature being actually *true* in the generating model. To this end we performed an extensive set of experiments varying various parameters such as the search method in the learning algorithms, the sizes of the datasets, and the bootstrap method. Third, we also examine the bootstrap as providing information to guide the induction process. We look at the increase in performance when the learning procedure is biased with information from the bootstrap estimates.

Our experiments, in Section 4, yield the following results, that to the best of our knowledge are unknown on the application of the bootstrap for establishing the likelihood that a particular feature is in the generating model:

1. The bootstrap estimates are quite cautious. Features induced with high confidence are rarely false positives.

2. The Markov neighborhood and partial ordering amongst variables features are more robust than the existence of an edges in a PDAG.

3. The conclusions that can be established on high confidence features are reliable even in cases where the data sets are small for the model being induced.

In Section 5 we examine how to use the bootstrap estimated to induce higher scoring networks. These results are still preliminary but encouraging nevertheless. Altogether, these results provide strong evidence for the bootstrap as an appropriate method for extracting qualitative information about the domain of study from features in the induced Bayesian network.

The study of methods for establishing the quality of induced Bayesian networks has not been totally ignored in the literature. Cowell et al. [5] present a method based on the log-loss scoring function to *monitor* each variable in a given network. These monitors check the deviation of the predictions by these variables from the observations in the data. Heckerman et al. [14] present an approach, based on Bayesian considerations, to establish the belief that a causal edge is part of the underlying generating model. The problem of confidence estimation that we study in this paper, is similar in spirit to the one investigated by Heckerman et al. Yet, the basis of the approach and the algorithmic implementation is completely different. The relation is further explored in [10] where we propose (and show results) how the Bootstrap can be used to implement a "practical" Bayesian estimate of the confidence on features of models. For completeness we summarized this relation in Section 6.

## 2    Learning Bayesian Networks

We briefly review learning of Bayesian networks from data. For a more complete exposition we refer the reader to [12].

Consider a finite set $\mathbf{X} = \{X_1, \ldots, X_n\}$ of discrete random variables where each variable $X_i$ may take on values from a finite set. We use capital letters, such as $X, Y, Z$, for variable names and lowercase letters $x, y, z$ to denote specific values taken by those variables. Sets of variables are denoted by boldface capital letters $\mathbf{X}, \mathbf{Y}, \mathbf{Z}$, and assignments of values to the variables in these sets are denoted by boldface lowercase letters $\mathbf{x}, \mathbf{y}, \mathbf{z}$.

A *Bayesian network* is an annotated directed acyclic graph that encodes a joint probability distribution of a set of random variables $\mathbf{X}$. Formally, a Bayesian network for $\mathbf{X}$ is a pair $B = \langle G, \Theta \rangle$. The first component, namely $G$, is a directed acyclic graph whose vertices correspond to the random variables $X_1, \ldots, X_n$, and whose edges represent direct dependencies between the variables. The graph $G$ encodes the following set of independence statements: each variable $X_i$ is independent of its non-descendants given its parents in $G$. The second component of the pair, namely $\Theta$, represents the set of parameters that quantifies the network. It contains a parameter $\theta_{x_i \mid \mathbf{pa}(x_i)} = P_B(x_i \mid \mathbf{pa}(x_i))$ for each possible value $x_i$ of $X_i$, and $\mathbf{pa}(x_i)$ of $\mathbf{pa}(X_i)$, where $\mathbf{pa}(X_i)$ denotes the set of parents of $X_i$ in $G$. A Bayesian network $B$ defines a unique joint probability distribution over $\mathbf{X}$ given by:

$$P_B(X_1, \ldots, X_n) = \prod_{i=1}^{n} P_B(X_i \mid \mathbf{pa}(X_i)).$$

The problem of learning a Bayesian network structure can be stated as follows. Given a *training set* $D =$

$\{\mathbf{x}[1], \ldots, \mathbf{x}[N]\}$ of instances of $\mathbf{X}$, find a network $B$ that *best matches* $D$. The common approach to this problem is to introduce a scoring function (or a *score*) that evaluates the "fitness" of networks with respect to the training data, and then to search for the best network (according to this score). In this paper we use the score proposed in [13] which is based on Bayesian considerations, and which scores a network structure according to the posterior probability of the graph structure given the training data (up to a constant).

We note that the derivation of such score treats the problem as a density estimation problems. The desire is to construct networks that will assign high probability to new (previously unseen) data from the same source. The structural features of the networks are induced indirectly, since presumably the "right" structure is the one that can better generalize from the training data.

Finding the structure that maximizes the score is usually an intractable problem [4]. Thus, we usually resort to heuristic search to find a high-scoring structure. Standard proposals for such search include greedy hill-climbing, stochastic hill-climbing, and simulated annealing; see [13]. In this paper, we will use a greedy hill-climbing strategy augmented with TABU lists and random restarts to escape local maxima.

In our experiments, we will not assess directly the confidence on the features of the induced network, but rather, on the features in the class of networks that are equivalent to it. Two Bayesian network structures $G$ and $G'$ are *equivalent*, if they imply exactly the same set of independence statements. The characterization of Bayesian network equivalence classes is studied in [3, 18, 19, 20]. Results in these papers establish that equivalent networks agree on the connectivity between variables, but might disagree on the direction of the arcs. These results also show that each equivalence class of network structures can be represented by a *partially directed graph* (PDAG), where a directed $X \to Y$ denotes that all members of the equivalence class contain the arc $X \to Y$; and, an undirected edge $X\!-\!Y$ denotes that some members of the class contain the arc $X \to Y$, and some contain the arc $Y \to X$. The score in [13] is *structure equivalent* in the sense that equivalent networks receive the same score. In our experiments, we learn network structures and then use the procedure described in [3] to convert them to to PDAGs.

## 3   Bootstrap for Confidence Estimation

Let $G$ be a network structure. A *feature* of interest in this structure might be the existence of an $X \to Y$ in the PDAG that corresponds to $G$. Another feature of interest might be that $X$ precedes $Y$ in the PDAG that corresponds to $G$. In general, we can treat these features as functions from network structures into the set $\{0, 1\}$. We will usually use the letters $f$ and $g$ to denote features.

Suppose we are given a data set of $N$ observations

$D = \{\mathbf{x}[1], \ldots, \mathbf{x}[N]\}$, each an assignment of values to $\mathbf{X}$. Moreover, assume that these assignments were sampled independently from a probabilistic network $B$ with structure $G$. Let $\hat{G}(D)$ be the network structure returned by our induction algorithm invoked with data $D$ as input. For any feature $f$ consider the following quantity

$$p_N(f) = \Pr\{f(\hat{G}(D)) = 1 \mid |D| = N\}.$$

This is the probability of inducing a network with the feature $f$ among all possible datasets of size $N$ that can be sampled from $B$.[1] If our induction procedure is *consistent*, then we expect that as $N$ grows larger, $p_N(f)$ will converge to $f(G)$. That is, we will give $f$ confidence close to one if it holds in $G$, and close to $0$ if it does not.

The quantity $p_N(f)$ is a natural measure of the power of any induction algorithm. Our goal is to estimate $p_N(f)$, given only a single set of observations $D$ of size $N$. This would mimic the usual induction situation when we want to learn a model from data. We now describe two possible algorithms: the parametric and non-parametric bootstraps.

We start with the non-parametric bootstrap. The underlying intuition is that we should be more confident on features that would still be induced when we "perturb" the data. The question is how to perturb the data and yet maintain the general statistical features of the dataset. In the non-parametric bootstrap we generate such perturbations by re-sampling from the given dataset. We then estimate confidence in a feature by examining in how many of the perturbed datasets it appears induced. The non-parametric bootstrap is performed by executing the following steps:

- For $i = 1, 2, \ldots m$
  - Re-sample, with replacement, $N$ instances from $D$. Denote by $D_i$ the resulting dataset.
  - Apply the learning procedure on $D_i$ to induce a network structure $\hat{G}_i = \hat{G}(D_i)$.
- For each feature of interest, define

$$p_N^{*,n}(f) = \frac{1}{m} \sum_{i=1}^{m} f(\hat{G}_i).$$

The parametric bootstrap is a similar process. Instead of re-sampling the data with replacement from the training data, we sample new datasets from the network we induce from $D$:

- Induce a network $B$ from $D$.
- For $i = 1, 2, \ldots m$
  - Sample $N$ instances from $B$. Denote by $D_i$ the resulting dataset.

---

[1] More generally, we can consider the joint distribution of several features. Of course, there are nontrivial relationships between confidence estimates for different features. For example, if we consider edges in PDAGs, then clearly $p_N(X \to Y) + p_N(Y \to X) + p_N(X\!-\!Y) \le 1$.

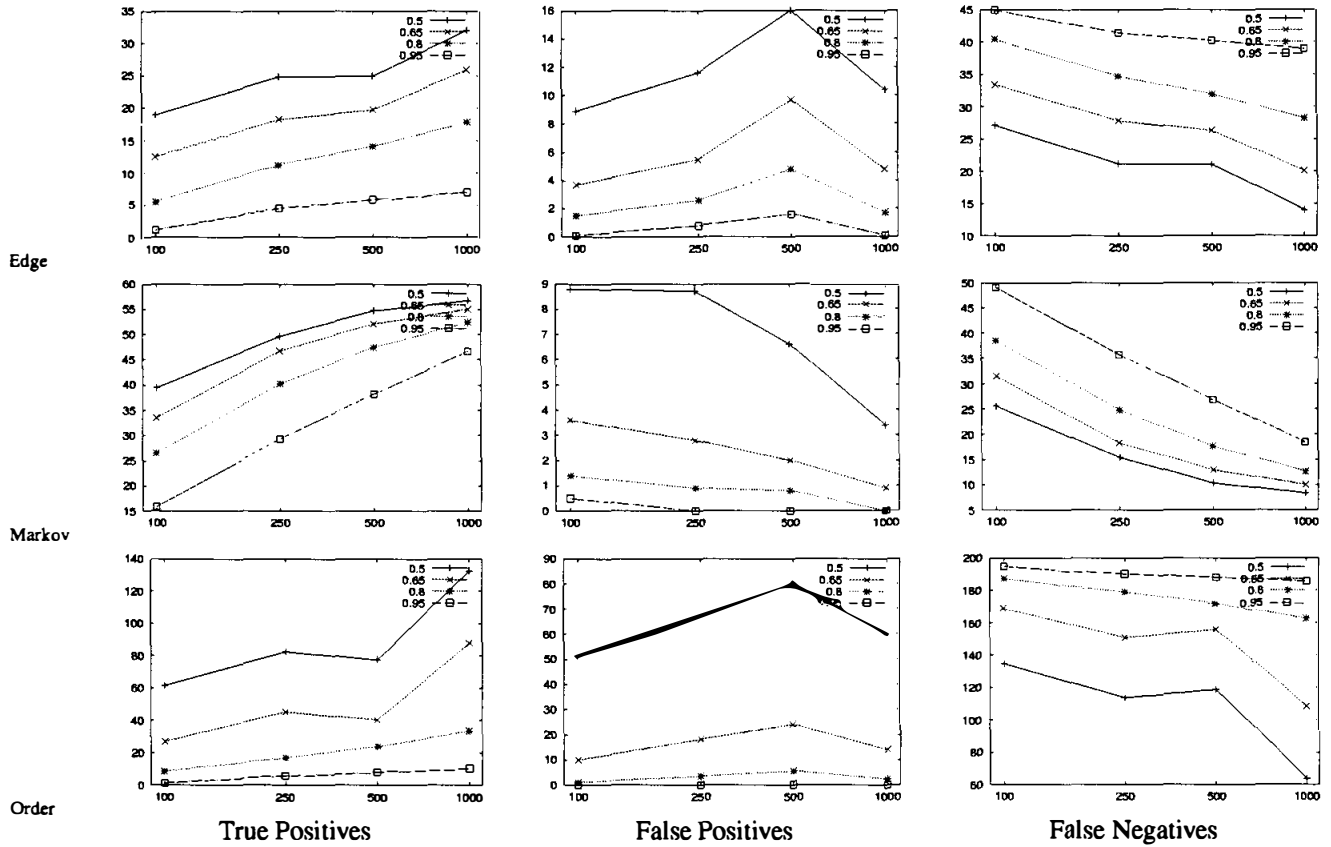

Figure 1: Quality of prediction of partially directed edges, markov neighborhoods, and orders in the alarm domain with non-parametric bootstrap. The columns correspond to average number of True Positives, False Positives, and False Negatives classifications. Each curve correspond to a value of the confidence threshold $t$. The $x$-axis shows the number of instances, and the $y$-axis shows the average number of edge features in each category. These averages are taken from bootstrap estimates, each with 100 resamples, from 10 datasets sampled from the "alarm" network.

- Apply the learning procedure on $D_i$ to induce a network structure $\hat{G}_i = \hat{G}(D_i)$.

- For each feature of interest, define

$$p_N^{*,p}(f) = \frac{1}{m} \sum_{i=1}^{m} f(\hat{G}_i).$$

The parametric bootstrap is quite different than the non-parametric one in the following sense. We are using simulation to answer the question: If the true network was indeed $B$, could we induce it from datasets of this size? By answering this question we can determine the level of confidence in the results of our induction.

We note that main computational cost in both variants of the bootstrap is dominated by the repeated calls to the induction procedure, and not by the since the sampling steps.

An important question is under what conditions will the Bootstrap estimate converge. Namely, under what conditions $|p_N(.) - p_N^*(.)|$ will approach 0 as $m$ and $n$ tend to $\infty$. The parametric bootstrap estimates of $p_N(e)$ will converge under more general conditions than the non-parametric bootstrap, provided, of course, that the parameterization converges to the true underlying model at least asymptotically. On the other hand, if this last condition is not satisfied then no consistency claim can be made. The non-parametric bootstrap estimates require no such model consistency. The consistency of the non-parametric bootstrap, however, requires uniform convergence in distribution of the bootstrap statistic as well as a continuity condition (in the parameters). The experiments and results presented in [10] were designed to verify convergence in both types of bootstrap for the features tested (existence of an edge in the PDAGS). We are currently working on providing a thorough theoretical analysis of these conditions in the context of Bayesian network induction. The experiments in the next section test to what extend we can use the bootstrap estimates as expressing the likelihood that the features tested belong to the generating model.

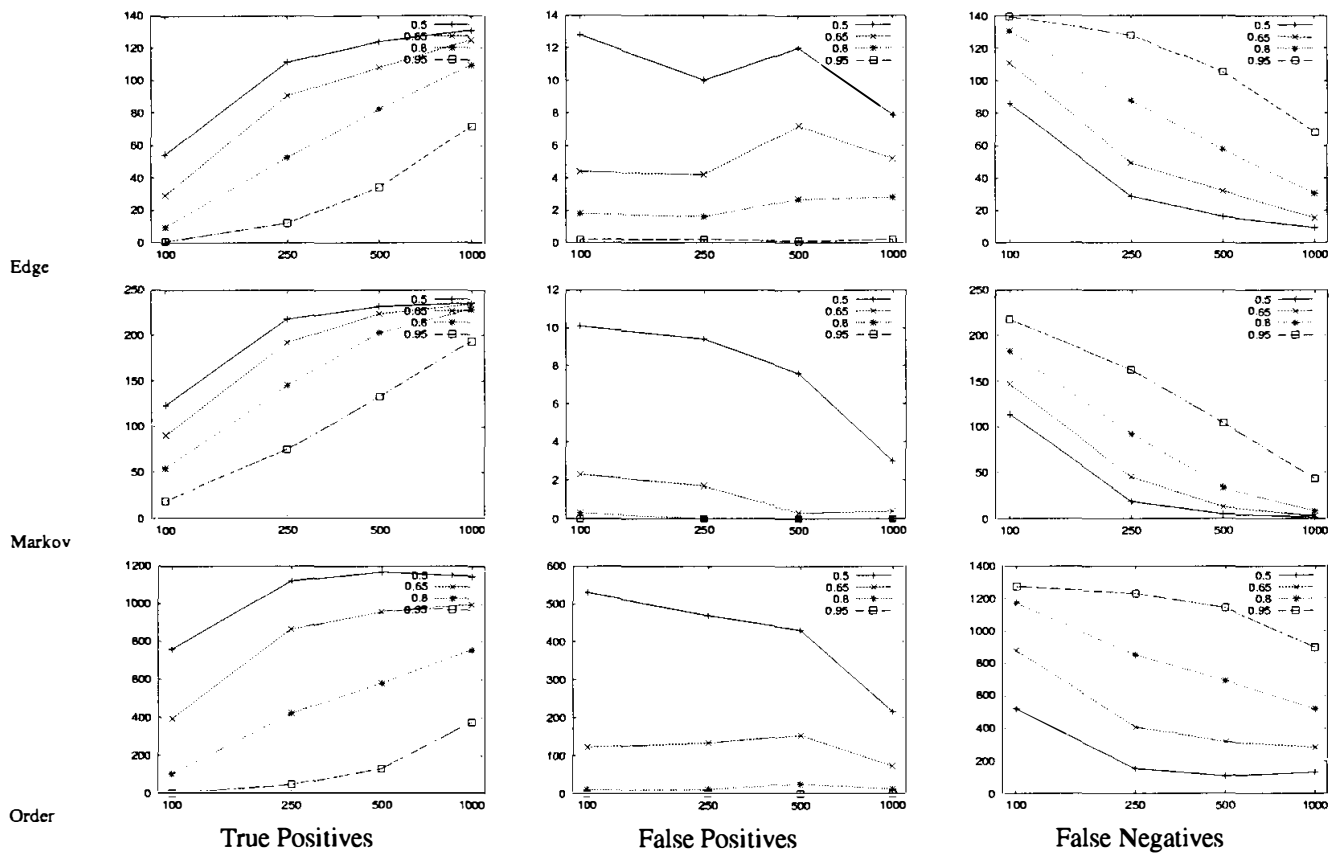

Figure 2: Quality of prediction of partially directed edges, markov neighborhoods, and orders in the gene domain with non-parametric bootstrap. (See caption of Figure 1 for details.)

## 4    Empirical Evaluation

To test the bootstrap, we use synthetic data that we generated from known models. This allows us to compare the features that the bootstrap is confident about to the true features in the generating network. Thus, for example, if our bootstrap confidence on node $X$ belonging to the Markov blanket of node $Y$ is high (above a determined threshold), we expect $X$ to be in the Markov blanket of $Y$ in the generating model. In addition, we also want to characterize how does the bootstrap estimates depend on various parameters, such as size of dataset, type of feature, and bootstrap method.

### 4.1    Methodology

We performed simulation results from three networks:

- alarm [1]. This network has 37 random variables and 46 edges, only 4 of which are undirected in the PDAG. This is a standard benchmark in the learning literature.

- gene. A network induced using a gene expression dataset from [8] for 76 genes. Genes were grouped by a clustering algorithm that searches for groups of related genes (details of the induction can be found in

[11]). The network has 140 edges, only 5 of which are undirected in the PDAG.

- text. A network induced from a dataset of messages from 20 newsgroup [15]. Each document is represented as an instance with a variable denoting the newsgroup, and 99 boolean variables corresponding to most frequent words (other than stop words) and denoting whether the word appears in the message. The network has 350 edges, only 12 of which are undirected in the PDAG.

¿From these networks, we performed experiments with $N$ (the number of instances in our data set) being $100, 250, 500, 1,000$. For each network and sample size, we sampled 10 "input" datasets for the bootstrap procedure. We then applied both the parametric and non-parametric bootstraps with $m = 100$.

In all of our experiments, we used the BDe score of [13] with a uniform prior distribution with equivalent sample size 5. This prior was chosen as a relatively uninformative one. The search procedure we used is a greedy hill-climbing search with random restarts. This procedure attempts to apply the best scoring change to the current network until no further improvement can be made. Once the hill-climbing procedure is stuck at a local maxima, it applies 20 random arc changes (addition/deletion/reversal) and restarts the search. The search is terminated after a fixed number of restarts.

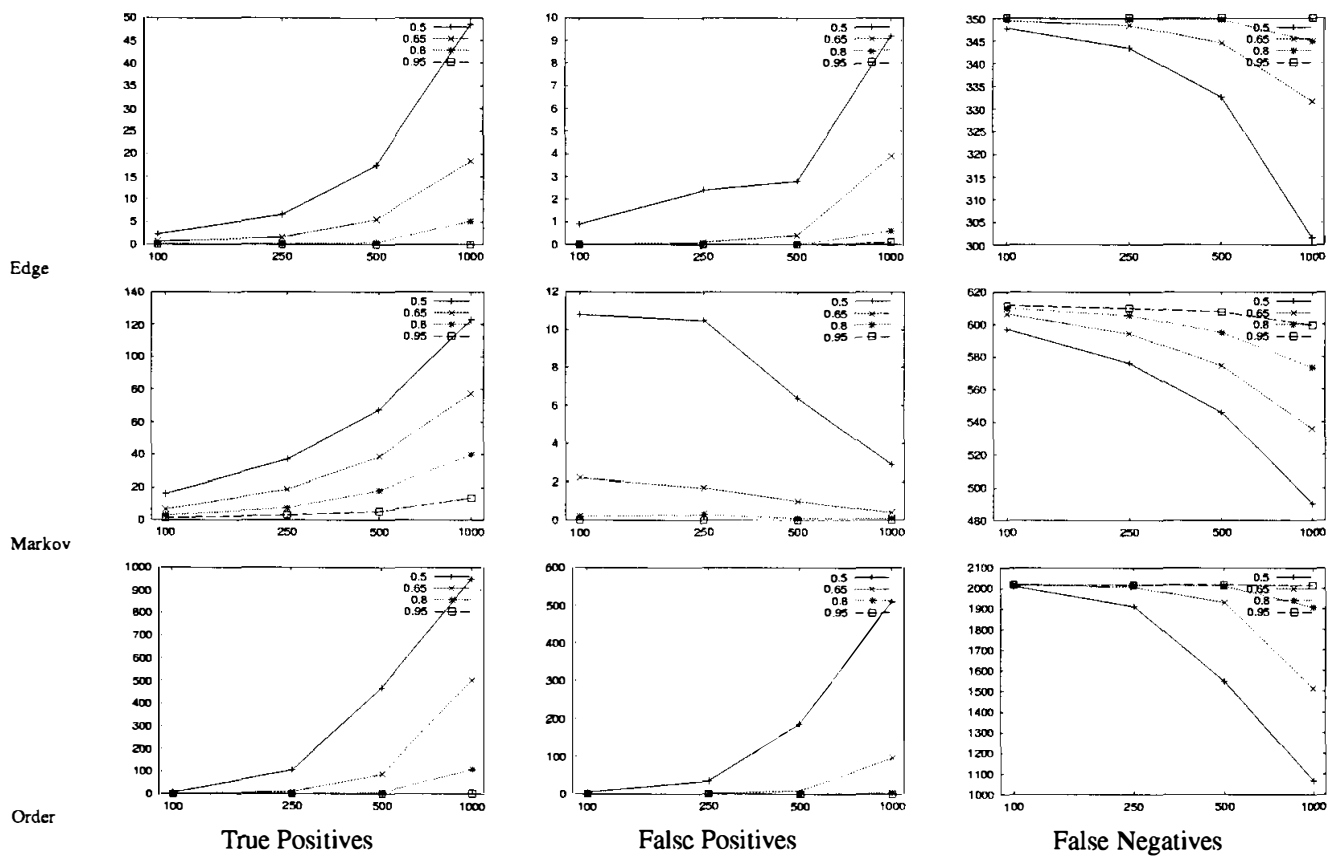

Figure 3: Quality of prediction of partially directed edges, markov neighborhoods, and orders in the text domain with non-parametric bootstrap. (See caption of Figure 1 for details.)

We computed the bootstrap estimates for three types of features:

- Edges in PDAGS. We treat the directed and undirected edges between pairs of variables as different features.

- Ordering relations of the form "$X$ is an ancestor of $Y$" in the PDAG.

- Markov neighborhoods, of the form "$X$ is in the Markov blanket of $Y$" (or vice verse). Two variables are Markov neighbors if there is an arc between them, or if they are both parents of another variable.

## 4.2 Evaluation

There are many possible ways of interpreting the bootstrap results. Perhaps, the simplest is to select a threshold $t$, and report all features that with $p_N(f) \geq t$. This way we can label all features as either "positive", if the confidence in them is above the threshold, or "negative", if it is below the threshold. Given such a labeling of features, we can measure the number of "true positives", correct features of the generating network that are correctly labeled, "false positives", wrong features that are labeled as positives, "false negatives", correct features that are labeled as negative, and "true negatives", wrong features that are labeled correctly. We report the numbers in the first three categories in prediction of the three type of features in Figures 1, 3, and 2, for the alarm, gene, and text domains respectively. The reported numbers are averaged over the estimates generated by the 10 non-parametric bootstrap runs.

There are several noticeable trends in these results. First, as expected, as the number of instances grow, the prediction quality improves. That is, the number of true positives increases, and the number of false positives and false negatives decreases. In addition, since as we increase the threshold we label fewer features as positive, the number of true positives and false positives decreases, while the number false negative increases.

Second, and more interestingly, the bootstrap samples are quite cautious. As we can see, the number of false positives is usually smaller than the number of true positives and false negatives. (Note the different scales in the graphs.) Thus, most of the prediction errors are one-sided in that they usually omit correct features and do not include incorrect ones.

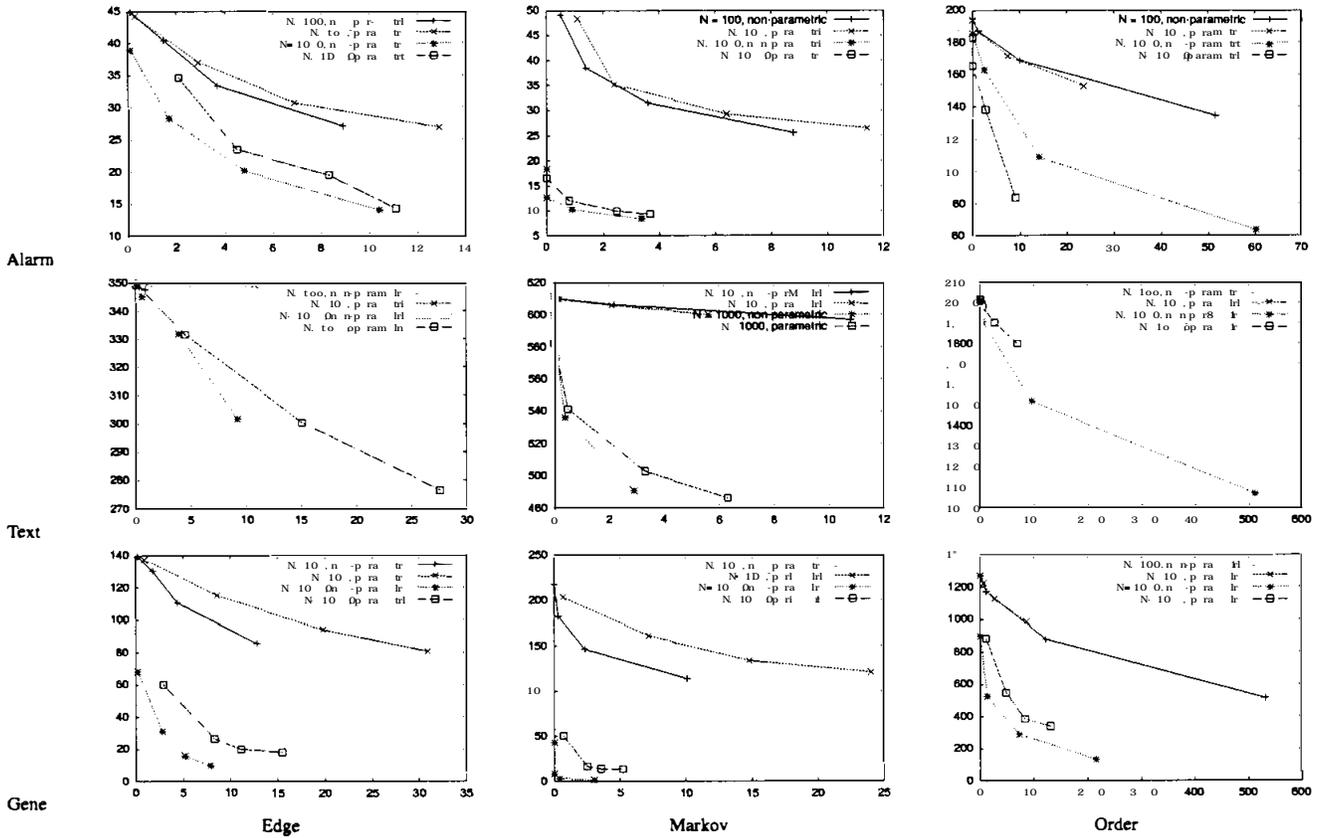

Figure 4: Comparison of parametric to non-parameteric bootstrap. The $x$-axis shows the average number of false positives and the $y$-axis shows the average number of false negatives. The curve shows the tradeoff between false positives and false negatives at different values of $t$, $0.95$, $0.8$, $0.75$, and $0.5$, for the predictions of a particular feature by one of the procedures. The columns correspond to the type of feature predicted.

Third, we notice that the "reasonable" level of confidence for thresholding depends on the domain. For example, in the alarm domain setting $t = 0.8$ leads to few false positives and a reasonable number of true positives. On the other hand, in the text domain, setting $t = 0.8$ leads few positives predictions, while setting $t = 0.65$ returns few false positives. Thus, we might be inclined to use this lower threshold value in this domain. It is unclear to us at this stage what is the source of this phenomena.

Finally, some features are easier to predict than others. For example, the prediction of Markov neighborhood of two variables is more robust than that of PDAG edges. Similarly, ordering information can also be quite reliably predicted based on the bootstrap confidence measures. This last observation is a bit surprising. Clearly, the "long-range" orderings between variables are a function of edge direction. Thus, the fact that we can predict some of them reliably indicates that some variables are recognized as ancestors of others, although this relation is determined by different directed paths in different bootstrap runs.

The ability to predict Markov neighborhoods, on the other hand, seem in line with common sense. This type of feature is less sensitive to the exact ordering between variables. In fact, it might be argued that these features might be easily estimated by other methods. To test this, we performed a simple test (suggested by an anonymous reviewer): instead of learning networks in the bootstrap samples, we learned Bayesian networks with in-degree at most one. These networks are easy to learn and take into account only pairwise interactions between variables. Figure 5 shows the tradeoff curves for non-parametric bootstrap using networks and trees. As we can see, the tree-based estimates are worse (both in terms of false positives and false negatives), except for the text domain. We suspect that this is partially due to the sparse nature of the source network in this domain.

As a conclusion, the bootstrap confidence measures are quite informative about the generating distribution. Moreover, some global features, such as partial ordering relations, can be determined from small data sets.

Next, we compared the parametric bootstrap to the non-parametric one. Figure 4 shows graphs of false positives vs. false negative tradeoffs between the two methods. Although, the performance of the two methods is similar, these graphs suggest that non-parametric bootstrap has better performance. The performance curves for the non-parametric bootstrap are usually closer to the origin, implying a smaller number of errors.

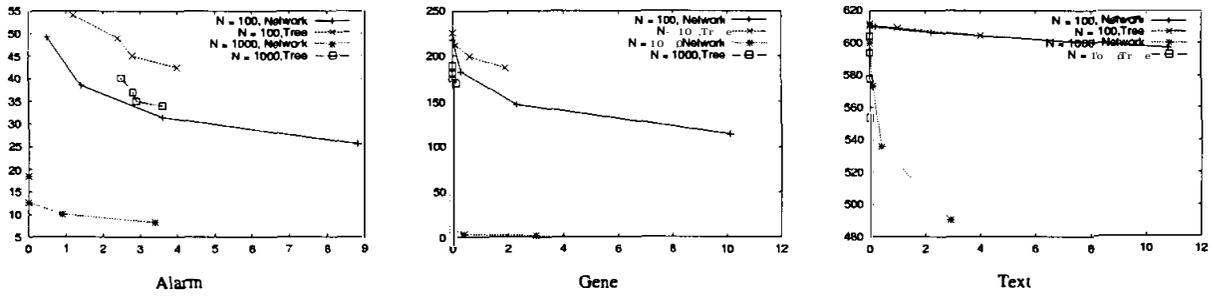

Figure 5: Comparison of non-parameteric bootstrap with network and with trees in prediction of Markov neighborhoods. (See caption of Figure 4 for details.)

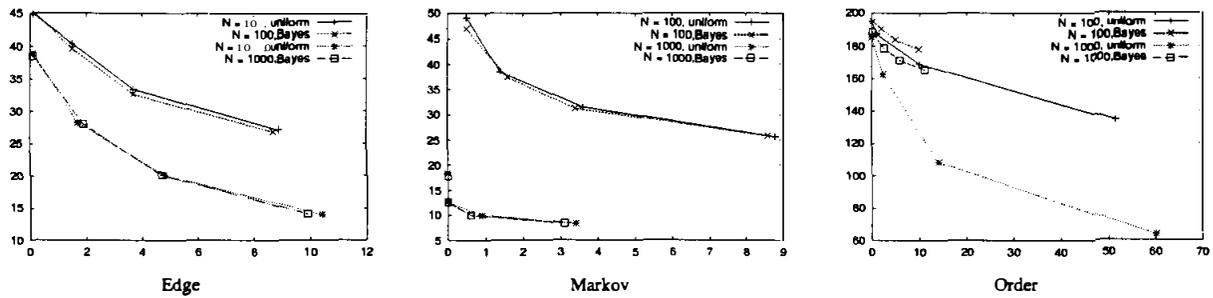

Figure 6: Comparison of non-parameteric bootstrap and Bayesian weighted non-parametric bootstrap on the alarm domain. (See caption of Figure 4 for details.)

| Domain | $N$ | Constrained | | Unconstrained | |
|---|---|---|---|---|---|
| | | avg. | s.d. | avg. | s.d. |
| alarm | 100 | -21.64 | 0.65 | -21.51 | 0.61 |
| | 250 | -18.39 | 0.36 | -18.41 | 0.37 |
| | 500 | -17.01 | 0.17 | -17.02 | 0.16 |
| | 1000 | -16.17 | 0.20 | -16.19 | 0.20 |
| gene | 100 | -58.81 | 2.03 | -58.85 | 2.10 |
| | 250 | -52.25 | 1.23 | -52.53 | 1.22 |
| | 500 | -48.04 | 0.62 | -48.33 | 0.67 |
| | 1000 | -45.21 | 0.52 | -45.70 | 0.67 |
| text | 100 | -60.44 | 1.00 | -60.44 | 1.00 |
| | 250 | -57.87 | 0.74 | -57.77 | 0.76 |
| | 500 | -56.06 | 0.69 | -55.90 | 0.57 |
| | 1000 | -54.68 | 0.52 | -54.69 | 0.53 |

Table 1: Average and standard deviation of normalized scores (BDe score divided by $N$) for networks learned with and without using the ordering constrained from the non-parametric bootstrap estimation.

| Domain | $N$ | Constrained | | Unconstrained | |
|---|---|---|---|---|---|
| | | avg. | s.d. | avg. | s.d. |
| alarm | 100 | -17.67 | 0.26 | -17.54 | 0.22 |
| | 250 | -16.04 | 0.16 | -16.05 | 0.18 |
| | 500 | -15.63 | 0.07 | -15.62 | 0.06 |
| | 1000 | -15.34 | 0.02 | -15.36 | 0.03 |
| gene | 100 | -54.73 | 1.18 | -54.82 | 1.02 |
| | 250 | -47.30 | 0.51 | -47.73 | 0.35 |
| | 1000 | -42.01 | 0.26 | -42.55 | 0.52 |
| text | 100 | -57.08 | 0.24 | -57.05 | 0.24 |
| | 250 | -55.57 | 0.07 | -55.57 | 0.07 |
| | 500 | -54.53 | 0.07 | -54.54 | 0.09 |
| | 1000 | -53.62 | 0.06 | -53.65 | 0.07 |

Table 2: Average and standard deviation of test set log-loss of the networks learned with and without using the constraints from the non-parametric bootstrap estimation.

## 5    Bootstrap for Network Induction

A common idea in learning is the use of prior knowledge. In particular, when learning structure, we can use prior knowledge on the structures we are searching to reduce the size of the search space, and thus improve both the speed of induction and more importantly, the quality of the learned network. Commonly used prior information include ordering constraints on the random variables, or the existence of certain arcs. In this section we explore the use of the Bootstrap for determining this information. The proposal consists of re-sampling from the dataset to induce bootstrap sample and then gather estimates on the confidence of these features. Then, we can use structural properties with high confidence to constrain the search process.

As a preliminary exploration of this idea, we performed the following experiment. We generated non-parametric bootstrap samples, and collected from them two types of constraints. First, if the estimate that $X$ precedes $Y$ has confidence higher than $0.8$, then we require that the learned network will respect this order. That is, we disallow learning networks where $Y$ is an ancestor of $X$. In addition, if the confidence that $X$ is in the Markov neighborhood of $Y$ is *smaller* than $0.05$, then we disallow $Y$ as a parent of $X$. The intuition, is that if $X$ and $Y$ are closely related, then we should be able to detect that in our bootstrap runs. If only a tiny fraction of the bootstrap networks have these two variables connected to each other, then they are probably not related.

After collecting these constraints, we invoke the search procedure to learn a network from the original data set, but we restrict it to consider only structures that satisfy the given constraints. We repeated this experiment 10 times for different initial data sets. In Table 1 we report the score of the networks induced by this procedure. In Table 2 we report the error from the generating distribution (measured in terms of log-likelihood assigned to test data) for the same networks.

These results show that for small training sets we can find slightly better scoring networks using the constraints generated by the bootstrap. Note that given the robustness of the estimates found in the previous section, these improvements can be trusted, even though in some cases the standard deviations of the scores and test set log-loss for the 10 experiments may seem relatively large. We should remember, however, that most of this variance is due to the small sample size.

## 6    Discussion: Bayesian estimation

The Bayesian perspective on confidence estimation is quite different than the "frequentist" measures we discussed above. A Bayesian would compute (or estimate) the posterior probability of each feature. Via reasoning by cases this is simply:

$$\Pr(f \mid D) = \sum_{G} \Pr(G \mid D) f(G). \qquad (1)$$

Where $f$ denotes the feature being investigated and the term $\Pr(G \mid D)$ is the posterior of a structure given the training data, and for certain classes of priors, can be computed up to a multiplicative constant (where the constant is the same for all graphs $G$) [13].

A serious obstacle in computing this posterior is that it requires summing over a large (potentially exponential) number of equivalence classes. Heckerman et al. [14] suggest to approximate (1) by finding a set $\mathcal{G}$ of high scoring structures, and then estimating the relative mass of the structures in $\mathcal{G}$ that contains $f$.

$$\Pr(f \mid D) \approx \frac{\sum_{G \in \mathcal{G}} \Pr(G \mid D) f(G)}{\sum_{G \in \mathcal{G}} \Pr(G \mid D)}.$$

This raises the question of how we construct $\mathcal{G}$. One simple approach for finding such a set is to record all the structures examined during the search, and return the high scoring ones. The set of structures found in this manner is quite sensitive to the search procedure we use. For example, if we use greedy hill-climbing, then the set of structures we will collect will all be quite similar. Such a restricted set of candidates also show up when we consider multiple restarts of greedy hill-climbing and beam-search. This is a serious problem since we run the risk of getting estimates of confidence that are based on a biased sample of structures. A way of avoiding this problem is to run an extensive MCMC simulation of the posterior of $G$. Then we might expect to get a more representative group of structures. This procedure, however, can be quite expensive in terms of computation time.

The bootstrap approach suggests a relatively cheap alternative. We can use the structures $\hat{G}(D_1), \ldots, \hat{G}(D_m)$ from the non-parametric bootstrap as our representative set of structures in the Bayesian approximation. In this proposal we use the re-sampling in the Bootstrap processes as way of widening the set candidates we examine. The confidence estimate is now quite similar to the non-parametric bootstrap, except that structures in the bootstrap samples are weighted in proportion to their posterior probability.

Figure 6 shows a comparison of the predictions of this approach with the non-parametric bootstrap on the alarm domain. The comparison on the other two domains is quite similar, so we omit it here. In general, the two approaches agree on high confidence features. This is not surprising, since the high confidence features appear in most of the bootstrap networks, and thus the Bayesian reweighting would still assign to them most of the mass. However, when we examine lower thresholds we can see some differences between the two approaches. This is particularly visible in the estimates of ordering relations.

We are currently exploring how to use the bootstrap in a more focused way to get a good approximation of the Bayesian posterior over features.

## 7 Conclusions

This paper proposes a methodology for computing confidence on features of an induced model based on Efron's bootstrap method. Whereas in a previous paper [10] we studied the bootstrap as assessing the degree of support of a *particular technique* towards a given feature, in this paper we examine the more important notion of confidence that assesses the likelihood that a given feature appears in the generating model. Our experiments lead to several conclusions: First, the bootstrap estimates are cautious and trustworthy; high confidence estimations seldom contain false positives. Second, features such as establishing a Markov neighborhood and partial ordering relations amongst variables are more robust than features such as the existence of an edges in a PDAG. Third, the conclusions that can be established on high confidence features are reliable even in cases where the data sets are small for the model being induced. These results extend, in our opinion, the role that adaptive Bayesian network are currently playing in data analysis tasks, enabling users to exploit the amount of qualitative information that the network structure provides about the domain.

We also provide preliminary results as of the use of the bootstrap for the induction of networks, and discussed its use in implementing a practical version of Bayesian estimates.

Finally, these results indicate that the bootstrap can be a reliable method for detecting latent causes. The problem of signaling the existence of latent causes, and uncovering the set of variables they should directly influence is of great interest. Given that we are computing estimates about the Markov blanket of each variables it would seems that a clique of variables that are definitely in each other's Markov blanket, but the edge relationships are unclear, would be indicative of the existence of a hidden variable. Given the reliability of these estimates we are optimistic about the results, and are currently experimenting with this approach.

### Acknowledgments

Some of this work was done while Nir Friedman and Abraham Wyner were at the University of California at Berkeley. We thank the NOW group at UC Berkeley and the MOSIX group at the Hebrew University for the use of their computational resources. Moises Goldszmidt was supported in part by DARPA's High Performance Knowledge Bases program under SPAWAR contract N66001-97-C-8548.